\pdfoutput=1

\documentclass[11pt]{article}

\usepackage[final]{acl}

\usepackage{times}
\usepackage{latexsym}
\usepackage{multirow}
\usepackage[T1]{fontenc}

\usepackage[utf8]{inputenc}

\usepackage{microtype}
\usepackage[inline]{enumitem}

\usepackage{inconsolata}

\usepackage{graphicx}

\usepackage{url}
\usepackage{diagbox}

\usepackage[breakable,skins]{tcolorbox}
\newtcolorbox{promptbox}[1][]{
    breakable,
    colback=gray!10,    
    colframe=gray!20,     
    title=#1,           
    fonttitle=\bfseries,
    boxrule=0.5mm,      
    arc=2mm,            
    outer arc=2mm,      
    coltitle=black,     
    enhanced,
}

\title{Empowering Healthcare Practitioners with Language Models:\\Structuring Speech Transcripts in Two Real-World Clinical Applications}

\author{
 \textbf{Jean-Philippe Corbeil},
 \textbf{Asma Ben Abacha},
 \textbf{George Michalopoulos},
\\
 \textbf{Phillip Swazinna},
 \textbf{Miguel Del-Agua},
 \textbf{Jérôme Tremblay},
 \textbf{Akila Jeeson Daniel},
\\
 \textbf{Cari Bader},
 \textbf{Yu-Cheng Cho},
 \textbf{Pooja Krishnan},
 \textbf{Nathan Bodenstab},
\\
 \textbf{Thomas Lin},
 \textbf{Wenxuan Teng},
 \textbf{Francois Beaulieu},
 \textbf{Paul Vozila}
\\
 Microsoft Healthcare \& Life Sciences
\\
 \small{
   \textbf{Correspondence:} \href{mailto:jcorbeil@microsoft.com}{\{jcorbeil,abenabacha,georgemi\}@microsoft.com}
 }
}

\begin{document}
\maketitle

\begin{abstract}
Large language models (LLMs) such as GPT-4o and o1 have demonstrated strong performance on clinical natural language processing (NLP) tasks across multiple medical benchmarks. Nonetheless, two high-impact NLP tasks — structured tabular reporting from nurse dictations and medical order extraction from doctor-patient consultations — remain underexplored due to data scarcity and sensitivity, despite active industry efforts. Practical solutions to these real-world clinical tasks can significantly reduce the documentation burden on healthcare providers, allowing greater focus on patient care. In this paper, we investigate these two challenging tasks using private and open-source clinical datasets, evaluating the performance of both open- and closed-weight LLMs, and analyzing their respective strengths and limitations. Furthermore, we propose an agentic pipeline for generating realistic, non-sensitive nurse dictations, enabling structured extraction of clinical observations. To support further research in both areas, we release SYNUR and SIMORD, the first open-source datasets for nurse observation extraction and medical order extraction.
\end{abstract}

\section{Introduction}

Large language models (LLMs), such as GPT-4o \cite{achiam2023gpt}, have rapidly advanced the state of the art across a broad range of natural language processing (NLP) tasks \cite{nori2023can, nori2024medprompt, abacha2024medec}, including in the clinical domain. Benchmarks like MedQA \cite{jin2021medqa}, PubMedQA \cite{jin2019pubmedqa}, and MMLU-Medical \cite{hendrycks2020mmlu} have shown that LLMs can achieve strong performance on standardized tasks. However, these benchmarks only scratch the surface of the real-world needs of clinical practice.

Two high-impact clinical NLP tasks remain underexplored: 1) structured reporting from nursing dictation, and 2) extraction of medical orders from long doctor-patient consultations. Both tasks are critically important in clinical workflows, where artificial intelligence can significantly reduce the documentation burden on healthcare providers and improve time-to-treatment for patients. Despite growing industry interest in this space, the lack of publicly available datasets and the inherent sensitivity of clinical data have limited progress and evaluation.

In this work, we examine the feasibility of LLM-based solutions for two key clinical information extraction tasks, leveraging both proprietary and newly open-source datasets. We provide a practical evaluation of their strengths and limitations using open- and closed-source LLMs, and highlight challenges such as handling long contexts, overflowing flowsheets, interpreting written representations of spoken language, and extracting complex structures. Furthermore, we introduce an agentic data generation pipeline that produces realistic, non-sensitive synthetic data for the nursing observation-extraction task, releasing the first open-source synthetic nursing dataset. Finally, we present the first medical order-extraction dataset, created through new annotations of existing doctor-patient conversations. These datasets offer valuable, realistic resources to complement existing clinical benchmarks and support further research.

\begin{figure*}[!h]
\centering
\includegraphics[width=1.05\linewidth, trim={0.7cm 0cm 0cm 0cm}, clip]{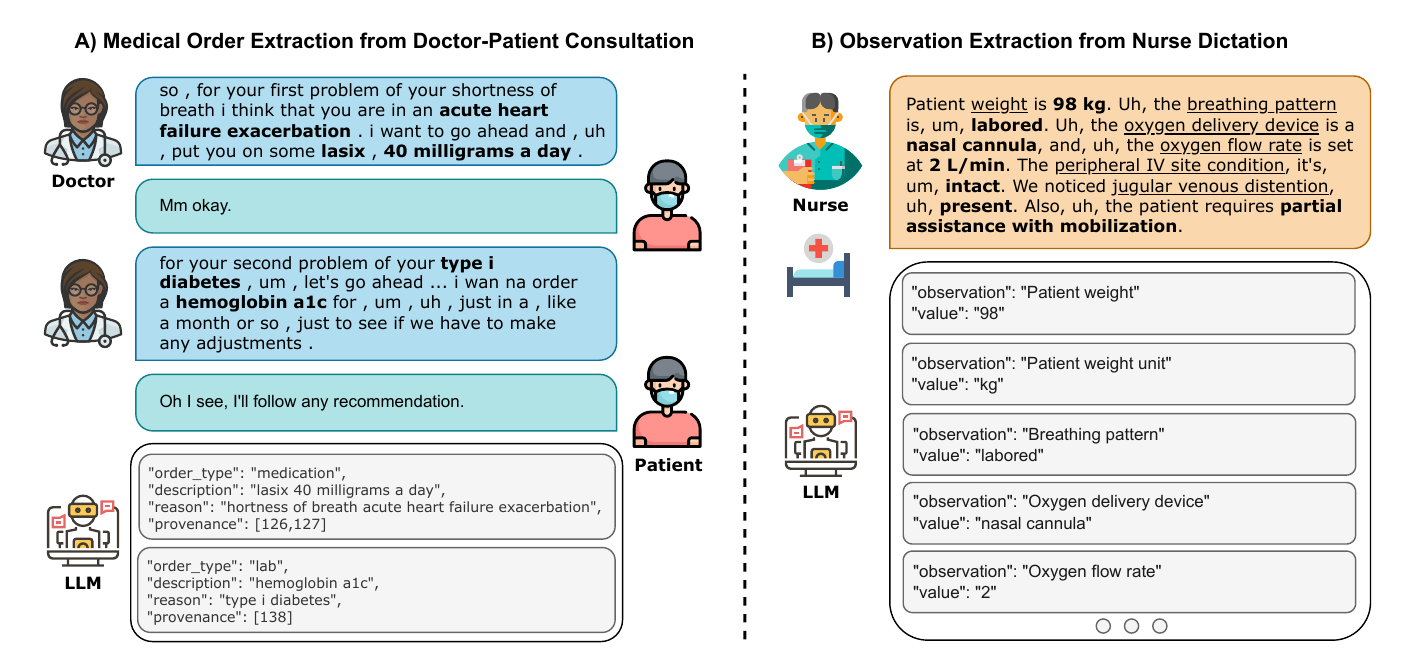}
\caption{Examples of our two real-world unstructured-to-structured tasks in clinical settings on spoken transcriptions from healthcare practitioners: A) medical order extraction from doctor-patient dialog, and B) observation extraction from nurse dictation. \small{\textit{Icon attributions in Appendix \ref{sec:icon}}}.}
\label{fig:diagram}
\vspace{-0.2in}
\end{figure*}

This paper makes four key contributions:
\begin{itemize}[topsep=0.1cm,itemsep=0cm]
    \item We release the \textbf{SYNUR} dataset, the first open-source SYnthetic NURsing dataset for extracting a wide range of structured observations from spoken nurse dictations\footnote{\url{https://hf.co/datasets/microsoft/SYNUR}}.
    \item We release the \textbf{SIMORD} dataset, the first open-source SIMulated ORDer dataset for extracting medical orders from doctor-patient simulated transcripts\footnote{\url{https://hf.co/datasets/microsoft/SIMORD}}.
    \item On SIMORD, we show that the 3.8B-parameter MediPhi-Instruct attains parity with GPT-4o (two-shot vs. one-shot) and surpasses it on the \emph{description} and \emph{reason} metrics, demonstrating the viability of lightweight open-weight models for this task.
    \item Using SYNUR and proprietary nursing datasets, we conduct the first systematic study of nursing observation extraction, highlighting challenges posed by long, disfluent dialogues and overflowing flowsheet slots.
\end{itemize}

\section{Previous Work}

\subsection{Nursing}



Speech interfaces promise faster, more complete nursing documentation but face challenges such as noise, privacy, and workflow disruption \cite{dinari2023nursespeech}. Surveys show growing interest in NLP for both nursing notes \cite{mitha2023nursingnote} and wider nursing tasks \cite{panchal2024nursingnlp}. Yet nurses still chart mainly in electronic health record (EHR) flowsheets, which are large, multi-tab tables with hundreds of data rows. So, any automation must mesh with that structure.

LLMs can already translate clinical narratives into structured variables with little or no task-specific training \cite{ling2023improving,Dagdelen2024Structured}. Yet their edge over fine-tuned encoder models remains debated \cite{gutierrez2022thinking,ling2023domain}. Consequently, no existing system simultaneously (i) ingests nurse dictations, (ii) conditions on the local flowsheet context, and (iii) outputs EHR-ready observations.

We target this gap by evaluating closed-weight LLMs in zero- and few-shot settings, trading fine-tuning for adaptability. The result is a dictation-to-flowsheet pipeline that inserts structured observations directly into the nurse’s existing charting interface without altering established routines.

\begin{figure*}[!t]
\centering
\includegraphics[width=1.02\linewidth]{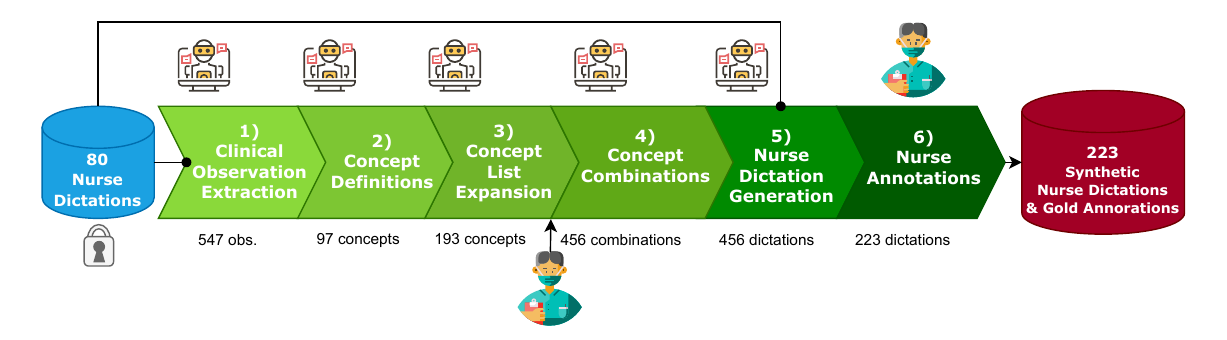}
\caption{SYNUR Dataset Creation Pipeline for generating realistic, synthetic nurse dictations with expert nurse annotations, comprising six steps: starting from 80 human-verified, fabricated dictations to final expert annotations. The number below each step represents the final amount of output elements generated at this specific step. The output of one step is the input of the next step, except for step 5 for which we also include 5 randomly sampled examples from the seed dictations. At step 6, the annotators also have access to the schema of step 3 along the synthetic dictation to produce its gold annotations.}
\vspace{-0.2in}
\label{fig:nurse_gen}
\end{figure*}

\subsection{Medical Order Extraction}
Order extraction is typically framed as a combination of named entity recognition and relation extraction. Early approaches primarily relied on rule-based systems such as MedEx \cite{medex} or support vector machines \cite{doan2010}. With the advent of contextual word embeddings, fine-tuned encoder models based on pre-trained transformer models such as BERT \cite{bert} and ClinicalBert \cite{alsentzer-etal-2019-publicly} have become the standard, achieving significant performance gains \cite{yang2020clinical,fabacher2025}. More recently, the emergence of LLMs has led to a paradigm shift in how these tasks are approached. Instead of span tagging, LLMs enable reformulating the task as a form of reading comprehension. Prompting techniques \cite{peng2023clinical, cui2023medtem2,PENG2024104630} have demonstrated strong performance on benchmark datasets, including adverse drug events \cite{henry20202018} and social determinants of health \cite{lybarger2023n2c2}. However, extracting orders from long patient–doctor dialogues is largely underexplored, especially the challenge of generating free-form fields (e.g. "description" and "reason") alongside structured ones (e.g. "order type").

\section{Data Creation Methods}

\subsection{Nursing Datasets}

\subsubsection{Proprietary Nursing Datasets}
Hospital-specific flowsheet schemas contain a list of all rows (keys) in the flowsheet (e.g., pulse rate), the data type associated with the row (e.g. numeric, picklist, etc.), and a list of possible values for picklist rows. The flowsheet schema can include hundreds to thousands of rows and does not include any data specific to any individual patient. In this work, we use the schemas of three different partner hospitals and for each hospital we also use an internal test set. These three test sets total 149 transcripts of rounding remarks dictated by nurses, annotated with 1,788 observations in aggregate.

Finally, we leverage available multishot examples to more explicitly connect rows with their possible verbalizations at extraction time. Each example consists of a transcript and its gold-truth row(s) corresponding to extractable medical observations. We organize examples into a database for each hospital schema and aim to select the best examples at runtime for a given transcript in a retrieval-augmented generation (RAG) setting \cite{lewis2020rag}.

\subsubsection{SYNUR: Synthetic Nursing Dataset}

We developed SYNUR, a synthetic nursing dataset, using 80 proprietary, fabricated seed dictations that were reviewed by practicing nurses for realism. Figure \ref{fig:nurse_gen} summarizes the six-stage pipeline, which alternates between the domain experts and the \textit{GPT-4o-0806} language model. We also provide all five prompts in Appendix \ref{sec:synur_prompts}.

Accurate nurse-observation extraction relies on comprehensive ontologies such as LOINC \cite{vreeman2018loinc,mcknight2019effective,loinc2025}, yet our seed notes did not align properly with LOINC. We therefore created an ontology before synthesizing new data:

\begin{enumerate}
    \item \textbf{Observation mining}: The LLM iteratively extracted observation phrases (e.g., “dark yellow”) and linked them to clinical concepts (e.g., “urine colour”) from the 80 notes, yielding 547 unique observations.
    \item \textbf{Concept consolidation}: We distilled these observations into 97 clinical concepts and assigned each a data type among boolean, integer, string, single-choice, or multi-choice.
    \item \textbf{Ontology expansion}: Leveraging the model’s medical knowledge, we asked it to propose additional concepts with example phrasings, expanding the ontology to 193 concepts, which an experienced nurse then validated and corrected.
    \item \textbf{Scenario generation}: We prompted the LLM with the final concept set to craft coherent patient-case rationale and compatible observation combinations.
    \item \textbf{Dictation synthesis}: Each scenario along with 5 randomly sampled seed dictations are passed to the LLM to generate a realistic nurse dictation that includes natural speech disfluencies and on-the-fly corrections.
    \item \textbf{Gold-standard labeling}: Expert nurse annotators verified the synthetic transcripts, and produced reference annotations using the ontology build at step 3 for a total of 223 dictations that contain 3000 observations.
\end{enumerate}

The same expert annotators labeled the real hospital sets and SYNUR.

\subsubsection{Quantitative Assessments of SYNUR}

\paragraph{Statistics} We compare the statistics of SYNUR to the hospital datasets in Table \ref{table_data_nurse}. We note that SYNUR is considerably larger at 223 samples, and its dictations tend to be longer --- i.e., 185 tokens on average compared to a maximum of 88 tokens.

\begin{table}[h!]
\centering
\small
\caption{Nursing dataset statistics.}
\begin{tabular}{|l|c|c|c|c|}
\hline
 & \textbf{H1} & \textbf{H2} & \textbf{H3} & \textbf{SYNUR} \\
\hline
Dataset Size & $49$ & $50$  & $50$  &  223 \\
\hline
AVG length & 88  & 28 & 61 & 185 \\
\hline
\end{tabular}
\label{table_data_nurse}
\vspace{-0.15in}
\end{table}

\paragraph{LLM-as-a-Judge} In Table \ref{tab:synur_judge}, we compare our hospital datasets with SYNUR using LLM-as-a-judge \cite{zheng2023judging} based on two criteria: quality (1-4) and realism (1-4). We leverage GPT-5 for this assessment. We observe that all scores are close to 3 which is considered high quality and realistic. We provide the prompt in Appendix \ref{sec:synur_llmasajudge}.

\begin{table}[h]
    \centering
    \small
    \caption{Average LLM-as-a-Judge scores to compare SYNUR to the three hospital datasets.}
    \begin{tabular}{|l|c|c|c|c|}
        \hline
         & H1 & H2 & H3 & SYNUR \\
        \hline
        Quality & 2.98 & 2.98 & 3.08 & 3.14 \\
        \hline
        Realism & 3.35 & 3.44 & 3.34 & 2.88 \\
        \hline
        AVG & 3.16 & 3.21 & 3.21 & 3.01 \\
        \hline
    \end{tabular}
    \label{tab:synur_judge}
\vspace{-0.15in}
\end{table}

\paragraph{Inter-Annotator Agreement} We provided a random sample of 10\% of the SYNUR dataset to two of our expert annotators, and we measured an accuracy of 86.8\% with a Cohen’s kappa of $0.634$, which points at a significant agreement.

\subsection{Medical Order Dataset}
\subsubsection{Doctor-Patient Consultation Datasets}
The long-form doctor-patient conversations used for the order-extraction task are primarily drawn from two datasets: ACI-Bench \cite{acibench} and PriMock57 \cite{primock57}. The ACI-Bench corpus comprises 207 naturalistic conversations between physicians and patients, curated by domain experts to reflect real-world clinical interactions. Similarly, the PriMock57 dataset contains 57 mock doctor-patient dialogues, designed to simulate clinical scenarios in a controlled setting.
Recent works such as Notechat \cite{wang2023notechat} has introduced large-scale synthetic dialogue datasets. While this corpus is the largest, we excluded it due to the prevalence of low-quality dialogues we observed.

\subsubsection{SIMORD: Simulated Orders Dataset}
We asked medically trained annotators to produce the gold-standard medical orders for the high-quality conversations of Primock57 and ACI-Bench. Annotation guidelines instructed to assess every medical order of type medication, imaging, lab, or follow-up within the conversation the way a doctor would create them in the EHR. This was intended to replicate doctors' current process executed at the end of a patient encounter. We measured an inter-annotator agreement of $0.768$. We sampled 100 examples containing 255 medical orders across both data sources as a test set and kept the others as training set (64 samples) used for few-shot prompting. Using the same methodology, we constructed an additional test set, which was used in the MEDIQA-OE Shared Task to evaluate participants' systems~\cite{MEDIQA-OE-Task}.

\section{Experiments \& Results}

 \subsection{Nurse Observation Extraction}

\begin{figure}[!h]
\vspace{-0.2in}
\centering
\includegraphics[width=1.06\linewidth,trim={0.7cm 0.7cm 0 0},clip]{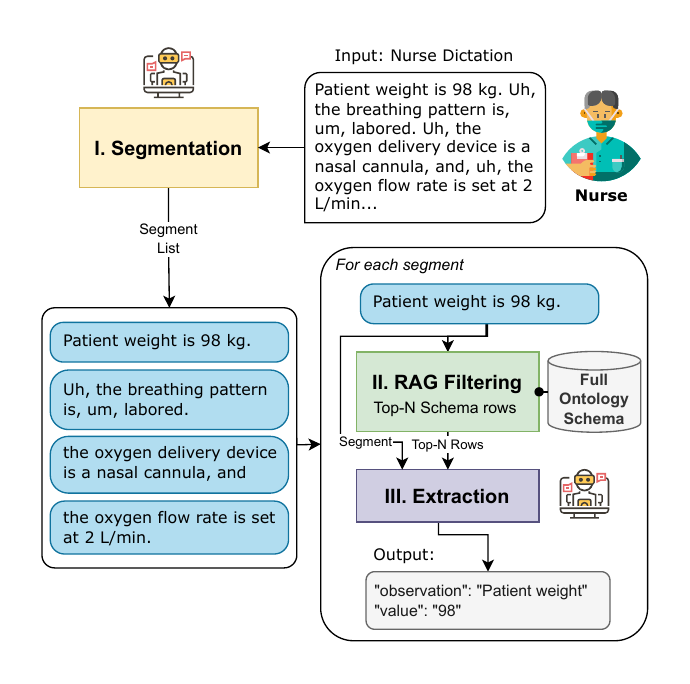}
\caption{Three-step method for the nurse observation extraction task.}
\vspace{-0.2in}
\label{fig:nurse_pipe}
\end{figure}

\begin{table*}[t!]
\centering
\caption{Nursing results across hospitals (H) and SYNUR for closed-weight LLMs. \small{Best values are in \textbf{bold}}.}
\vspace{-0.1in}
\begin{tabular}{|l|c|c|c|c|c|c|c|c|c|}
\hline
 & \multicolumn{5}{c|}{\small Zero-Shot ($f1$)} & \multicolumn{4}{c|}{\small Few-Shot ($f1$)}  \\
\hline
 \small{\diagbox[width=7em]{Model}{Dataset}} & \small{$\mathbf{SYNUR}$} & \small{$\mathbf{H_1}$} & \small{$\mathbf{H_2}$} & \small{$\mathbf{H_3}$} & \small{$\mathbf{H_{avg}}$} & \small{$\mathbf{H_1}$} & \small{$\mathbf{H_2}$} & \small{$\mathbf{H_3}$} & \small{$\mathbf{H_{avg}}$} \\
\hline
4o & \textbf{88.3} & 85.4 & 86.6 & 88.4 & 86.8 & 90.7 & 91.6 & 89.6 & 90.6 \\
4o-mini & 75.6 & 81.0 & 81.7 & 81.8 & 81.5 & 88.4 & 90.8 & 88.2 & 89.1 \\
4.1 & 86.7 & \textbf{88.2} & \textbf{87.6} & \textbf{89.6} & \textbf{88.5} & \textbf{91.2} & \textbf{93.6} & \textbf{92.4} & \textbf{92.4} \\
4.1-mini & 84.5 & 84.1 & 84.7 & 88.4 & 85.7 & 88.9 & 93.0 & 90.4 & 90.8 \\
\hline
\end{tabular}
\label{table_results_nurse}
\vspace{-0.15in}
\end{table*}
 
\subsubsection{Methodology}
The schema of our nursing approach is displayed in Figure \ref{fig:nurse_pipe}. It takes in a nurse's spoken assessment and then automatically populates the flowsheet by using an LLM to guide the extraction of key elements. Since flowsheets overflow most LLMs' context windows\footnote{With related issues such as lost-in-the-middle \cite{liu2024lost}.}, our approach contains three sub-tasks:
\begin{enumerate*}[label=(\roman*)] \item \textbf{Segmentation}: The segmentation step splits the streaming nurse transcript into medically coherent, 
continuous and non-overlapping segments via an LLM --- prompt in Appendix \ref{subs:Prompts}. \item \textbf{RAG}: We minimize the size of 
the schema by filtering potential rows to the top $N$ candidates given the current segment based on cosine similarity. \item \textbf{Extraction}: Finally, we use this prompt in Appendix \ref{subs:Prompts} with an LLM to extract the relevant text from the transcript segment and canonicalize it based on the (now reduced) schema. \end{enumerate*}

\subsubsection{Results}

To identify how different language models affect the performance of our nursing approach, we experimented with four closed-weight LLMs\footnote{Open-weight models were excluded because of compatibility considerations.}: \begin{enumerate*}[label=(\roman*)]    \item GPT-4o \item GPT-4o-mini \item  GPT-4.1   \item   GPT-4.1-mini  \end{enumerate*}. For all models we use the same prompt and we evaluate the ability of each model to correctly extract the flowsheet-relevant clinical facts from each transcript. In particular, we calculate the $f1$ score for each experiment. 


In Table \ref{table_results_nurse}, we can observe that the GPT-4.1 model consistently outperforms all other models across all proprietary datasets. However, it seems that GPT-4o achieved the best performance in the synthetic experiments, possibly because we use the same model to generate SYNUR. Furthermore, these experiments highlight the positive effect of including examples in the extraction prompt, as all models can achieve a higher $f_1$.
\paragraph{Error Analysis}
By examining misclassifications, we can identify two error categories. In particular, models struggle with \begin{enumerate*}[label=(\roman*)] \item identifying the value on a medical scale that corresponds to a given situation \item inferring a missing medical identifier \end{enumerate*}.

For SYNUR, we observe comparable performances except for 4o-mini for which the variations might be caused by the coarser-grained schema, longer dictations and variations in the speech style, e.g. more hesitations. Thus, future works could leverage real nurse dictations and build a large nursing observation ontology. Nonetheless, human verifications of SYNUR's schema and synthetic dictations mentioned high quality as well as high similarity to real data.

\subsection{Medical Order Extraction}
\subsubsection{Methodology}
The goal of the medical order-extraction task is to evaluate the performance of various language models in extracting structured orders from conversations between a patient and a physician. Each system is provided with a prompt that describes in detail the instructions to follow to generate a set of orders (see Appendix \ref{subs:Prompts}). Each order must contain the following four attributes:
i) \textbf{Description}: A concise summary of the order. ii) \textbf{Reason} (optional): The motivation behind the order. iii) \textbf{Type}: A categorical label indicating the nature of the order: medication, laboratory, follow-up, or imaging. iv) \textbf{Provenance}: The source or origin of the order. It is made up of a list of conversation line numbers where the order was prescribed. The output format is standardized as JSON, ensuring consistency and ease of downstream processing. 

We tested five closed-weight LLMs for this task: i) GPT-4o, OpenAI's multimodal flagship model \cite{achiam2023gpt}. ii) GPT-4.1, a refined version of GPT-4. iii) o1-mini, a lightweight reasoning model optimized for speed and cost. iv) o1-prev, the preview version of o1. v) o3-mini, a newer compact model with enhanced reasoning. Also, we experimented with open-weight language models including Phi3.5-mini-instruct (3.8B) \cite{abdin2024phi3} and its medical variant Mediphi-Instruct \cite{corbeil2024iryonlp} as well as Llama3-8B-instruct \cite{dubey2024llama3} and its medical variant Llama3-Med42-8B \cite{christophe2024med42v2}

Each model was prompted with identical inputs in both zero-shot and few-shot settings, and evaluated on the quality, completeness, and consistency of the extracted orders.

\subsubsection{Evaluation Metrics}

We evaluate model performance across five key metrics: \textbf{Match}, \textbf{Description}, \textbf{Reason}, \textbf{Type}, and \textbf{Provenance}. Results are reported after matching for each transcript reference and hypothesis orders based on description word overlap --- necessary to compare orders with each other. The \textbf{Match} score is computed from this alignment process as the F1 between reference and predicted orders without looking at the content, thus specifically accounting for the amount of fabricated or omitted orders. For \textbf{Description} and \textbf{Reason}, we compute unigram overlap F1 score using ROUGE \cite{lin-2004-rouge}. \textbf{Type} is evaluated using micro F1 score due to its finite set of discrete labels, and \textbf{Provenance} is assessed via F1 score over provenance indices\footnote{Turn numbers where the order originates in the transcript.}. The match score is the upper bound of the other metrics that are penalized by our precision/recall aggregation method accounting for empty values as demonstrated in Appendix \ref{sec:metric}.

\begin{table}[h!]
\centering
\setlength{\tabcolsep}{5pt}
\small
\caption{Results on SIMORD for open- and closed-weight LLMs with reasoning and few-shot variants. \textit{Desc} and \textit{Prov} are short for \textit{Description} and \textit{Provenance}, respectively. \small{Best values in \textbf{bold}}.}
\begin{tabular}{|l|l|c|c|c|c|c|}
\hline
& Model & \textbf{Match} &\textbf{Desc} & \textbf{Reason} & \textbf{Type} & \textbf{Prov} \\
\hline
& \multicolumn{6}{c|}{zero shot} \\
\hline
\multirow{4}{*}{\rotatebox{90}{open}} & Phi3.5 & 56.0 & 26.3 & 11.5 & 51.9 & 18.3 \\
& MP-Inst & 42.2 & 28.4 & 17.6 &  39.6& 11.2 \\
& Llama3 & 61.8 & 41.8 & 21.5 & 59.3 & 9.5 \\
& Med42 & 55.1 & 36.5 & 25.4 & 54.0 & 14.3 \\
\hline
\multirow{5}{*}{\rotatebox{90}{closed}} & 4o & 68.2 & 38.5 & 17.4 & 66.1 & 1.1 \\
& 4.1 & 67.0 & 30.7 & 12.1 & 65.2 & 1.1 \\
& o1-mini & 66.5 & 30.2 & 23.3 & 64.1 & \textbf{43.2} \\
& o1-prev & 55.0 & 28.1 & 17.0 & 53.8 & 13.4 \\
& o3-mini & 66.5 & 30.8 & 9.7 & 64.9 & 24.1 \\
\hline
& \multicolumn{6}{c|}{with one example} \\
\hline
\multirow{4}{*}{\rotatebox{90}{open}} & Phi3.5 & 64.5 & 41.4 & 20.7 & 60.6 & 8.6 \\
& MP-Inst & 61.0 & 45.7 & 32.6 & 59.3 & 9.8 \\
& Llama3 & 56.9 & 40.8 & 29.2 & 55.3 & 8.2 \\
& Med42 & 65.2 & 48.3 & 35.8 & 64.4 & 9.7 \\
\hline
\multirow{5}{*}{\rotatebox{90}{closed}} & 4o & 67.9 & 42.8 & 21.8 & 66.1 & 0.5 \\
& 4.1 & 67.6 & 33.0 & 12.7 & 66.0 & 1.2 \\
& o1-mini & 66.1 & 35.8 & 26.6 & 64.2 & 37.8 \\
& o1-prev & 65.2 & 40.2 & 24.6 & 63.5 & 14.8 \\
& o3-mini & \textbf{69.2} & 39.3 & 12.4 & \textbf{66.8} & 26.9 \\
\hline
& \multicolumn{6}{c|}{with two examples} \\
\hline
\multirow{4}{*}{\rotatebox{90}{open}} & Phi3.5 & 62.2 & 43.5 & 25.3 & 60.4 & 6.3 \\
& MP-Inst & 66.3 & \textbf{51.9} & 35.7 & 65.4 & 8.7 \\
& Llama3 & 62.8 & 47.6 & 30.8 & 61.5 & 4.5 \\
& Med42 & 63.2 & 49.2 & \textbf{37.9} & 62.4 & 9.6 \\
\hline
\end{tabular}
\label{table_results_oe}
\vspace{-0.2in}
\end{table}

\subsubsection{Results for Closed-Weight LLMs}

As shown in Table~\ref{table_results_oe}, no single closed-weight model consistently outperforms others across all metrics and settings. For instance, in the zero-shot setting, model \texttt{4o} achieves the highest score in \textbf{Description} (38.5\%) and \textbf{Type} (66.1\%), while \texttt{o1-mini} leads in \textbf{Provenance} (43.2\%). When examples are introduced, \texttt{4o} again performs best in \textbf{Description} (42.8\%), \texttt{o1-mini} in \textbf{Reason} (26.6\%), and \texttt{o3-mini} in \textbf{Type} (66.8\%). If we compare the \textbf{Type} scores to their upper-bound \textbf{Match} scores, we note that the gaps across closed-weight models are only of a couple percentage points indicating the ease of attributing a type. When we observe the \textbf{Description} scores in relation to the \textbf{Match} scores, we generally note a gap of more than 30\% suggesting room for improvement in \textbf{Description} generation. Similarly, the \textbf{Match} scores are also more than 30\% below a perfect extraction indicating the presence of fabricated or omitted orders. We observed one phenomenon that occurs occasionally --- especially for lab orders --- in which models tend to aggregate orders into one unique order leading to a perceived omission, because doctors often mentioned them sequentially in one dialogue turn.

\begin{table}[h!]
\centering
\caption{Average improvements on SIMORD for experiments on closed-weight models between zero-shot base models against one example or reasoning variants.}
\begin{tabular}{|l|c|c|c|c|}
\hline
\textbf{Metric} & \textbf{Desc} & \textbf{Reason} & \textbf{Type} & \textbf{Prov} \\
\hline
Example & 6.6\% & 3.8\% & 2.5\% & -0.3\% \\
Reasoning & -2.2\% & 2.9\% & -3.0\% & 25.7\% \\
\hline
\end{tabular}
\label{table_improvements_oe}
\end{table}

Table~\ref{table_improvements_oe} summarizes the average improvements across metrics when using examples or reasoning closed-weight models. Including a single randomly drawn example from the training set yields modest gains in \textbf{Description} (+6.6\%), \textbf{Reason} (+3.8\%), and \textbf{Type} (+2.5\%). This suggests that examples help models better contextualize and generalize, particularly for free-text fields. A dynamic RAG-based approach \cite{lewis2020rag} could likely further improve on this baseline.

Closed-weight reasoning models show a substantial improvement in \textbf{Provenance} (+25.7\%), indicating that this task benefits from deeper inference capabilities. However, these models do not show consistent advantages in other metrics, with slight declines in \textbf{Description} and \textbf{Type}, and only a marginal gain in \textbf{Reason}.

A notable practical issue is the variability in parsing reliability across models. Some models, particularly \texttt{o1-preview} and \texttt{GPT-4.1}, frequently produce malformed outputs --- especially due to run-on sequences in \textbf{Provenance} predictions --- which prevent successful JSON parsing. In contrast, models such as \texttt{4o}, \texttt{o1-mini}, and \texttt{o3-mini} maintain parsing error rates between 0-1\%, making them more reliable in deployment scenarios. As future work, constrained decoding \cite{willard2023guiding2,beurer2024guiding1} might fix such errors with caveats \cite{tam-etal-2024-speak}.


\subsubsection{Results for Open-Weight SLMs}



Table~\ref{table_results_oe} also presents the results for four open-weight language models using greedy decoding with a temperature 0 and a maximum of 1024 tokens. In the zero-shot setting, \texttt{Phi-3.5-mini-Instruct} (Phi3.5), \texttt{MediPhi-Instruct} (MP-Inst), and \texttt{Llama3-Med42-8B} (Med42) display limited overall performances. Nonetheless, Phi-3.5 records the highest \textbf{Provenance} accuracy among the open-weight systems, while \texttt{Llama-3-8B-Instruct} (Llama3) attains \textbf{Description} and \textbf{Reason} scores comparable to \texttt{GPT-4o} and \texttt{GPT-4.1}. Performance improves substantially with in-context learning. With two examples from the training set, MP-Inst achieves the top \textbf{Description} score surpassing \texttt{GPT-4o} and leads the open-weight group on the \textbf{Match} and \textbf{Type} metrics. Under the two-shot configuration, Llama3 attains the highest \textbf{Reason} score. 

Overall, we also notice parsing issues across open-weight models. While the zero-shot setting is affected at 10-20\%, the few-shot setting is more reliable at 1-4\%. Moreover, \textbf{Provenance} remains problematic for open-weight models also because of run-on sequences, highlighting the need for a multi-system solution.

Although comparing \texttt{GPT-4o} with one example to open-weight models with two examples is not strictly equivalent, the contrast remains informative given the large differences in parameter scale, model transparency, and inference latency.

\section{Conclusion}

We explored two clinically relevant NLP tasks: observation extraction from nurse dictations, and order extraction from doctor-patient conversations. Using both proprietary and open datasets, we evaluated the performances of open- and closed-weight LMs, identifying key strengths and bottlenecks. We introduced both SYNUR, the first synthetic nursing dataset, and SIMORD, the first medical order-extraction dataset. Our results demonstrate that LMs can reduce documentation burden in clinical workflows. These insights support the design and deployment of scalable, LLM-driven solutions in real-world healthcare settings.   



\section*{Limitations}

While SYNUR helps mitigate data scarcity and sensitivity issues in nurse dictations, synthetic data cannot fully capture the richness and variability of real clinical language. On the methodological side, the nursing codebase --- including its two core prompts --- cannot be shared in full; however, simplified versions of the segmentation and extraction prompts are provided in Appendix \ref{subs:Prompts}.

The SIMORD dataset is relatively small, with 64 training samples, which constrains its applicability for finetuning. Future work may address this by expanding the dataset or integrating synthetic examples. Although inter-annotator agreement is high, some variability remains in the annotations (e.g., span boundaries and informal conversational phrasing rather than formal writing). Such variation introduces natural noise that may bound achievable performance.

\bibliography{custom}

\appendix

\section{Appendix}
\label{sec:appendix}


\subsection{Prompts for Methods}
\label{subs:Prompts}

\begin{promptbox}[Simplified Nursing Segmentation Prompt]
\small
  Given an input TRANSCRIPT of a nurse's observations about a patient, your task is to divide the input TRANSCRIPT into contiguous SEGMENTS, based on clinical facts.
  
  A clinical fact refers to specific, verifiable information related to the health of a patient.
  
  TRANSCRIPT: 
  
  \%TRANSCRIPT\%
  
  SEGMENTS:
\end{promptbox}

\begin{promptbox}[Simplified Nursing Extraction Prompt]
\small

You are an expert at medical electronic health record flowsheet analysis.

Below is a TRANSCRIPT from a nurse dictation along with a flowsheet SCHEMA. Please extract the clinical observations from the TRANSCRIPT in strict, parsable JSON adhering to SCHEMA.

...

SCHEMA: \%SCHEMA\%

TRANSCRIPT: \%TRANSCRIPT\%

OUTPUT:
\end{promptbox}

\begin{promptbox}[Order Extraction Prompt]
\small
Act as if you are an experienced medical scribe. Based on the provided transcript, you must document the orders that are being
placed for the patient.

You must provide as output an array of json objects where each object is an order from the doctor.
Each order object should have keys for: description, order\_type, reason, provenance.

An example order json object is:

\{'description': 'CT of Chest',\\
'order\_type': 'imaging',\\
'reason':'infection',\\
'provenance': [45, 46]\}
     
The description is a textual description related to one specific order mentioned by the doctor.
It is the concatenation of doctor's text spans related to the order.
The order\_type options are "medication", "lab", "followup", "imaging".
The reason is one sentence explaining the diagnosis leading to that order.
The provenance is a list of line numbers where the different text spans in the order are extracted from.

The output must be a list of json objects. 
If there are no orders in the transcript, you should provide an empty array as output.
You absolutely need to limit your output to 4000 tokens.

===CURRENT ORDER EXTRACTION===

---DOCTOR TRANSCRIPT---
\end{promptbox}

\subsection{SYNUR Generation Prompts}
\label{sec:synur_prompts}
\begin{promptbox}[Step 1 - Clinical Observation Extraction]
\small
You are top entity extractor in clinical documents, very precise and very thorough in its work.

You will receive a speech description of a patient case as input. Your task is to extract a JSON array of fine-grained entity names.
You must extract the fine-grained concept names from the speech description in JSON array with JSON objects containing two keys: concept and text\_span.
You have a pool of encountered concepts from other speech description. You must use these concepts from the CONCEPT POOL or create a new concept that defines better an observation from a text span.
Examples of concepts are provided to make you understand the level of abstraction overall. It should be fine-grained, i.e. small groups of possible values encompass by one concept.

--EXAMPLE OF CONCEPTS--

Here are examples of concepts with potential observation forms.

\%EXAMPLES\%

--EXAMPLE JOB--

SPEECH DESCRIPTION

Patient abdomen is soft rounded and tenderness is present. All other body systems within defined limits. Blood pressure is 127/66.

OUTPUT

[

  \{
    "concept": "Abdomen exam",
    "span": "abdomen is soft rounded and tenderness is present"
  \},
  
  \{
    "concept": "General physical exam",
    "span": "All other body systems within defined limits"
  \},
  
  \{
    "concept": "Blood pressure",
    "span": "Blood pressure is 127/66"
  \}
  
]

--CONCEPT POOL--

\%CONCEPT\_POOL\%

--CURRENT JOB--

SPEECH DESCRIPTION

\%CONCEPT\_POOL\%

OUTPUT
\end{promptbox}

\begin{promptbox}[Step 2 - Concept Definitions]
\footnotesize
You are top clinical semantic expert in medical and clinical knowledge, very precise and very thorough in its work.
You will receive a concept with obsversations of this concept from real clinical documentations.
You task is to clean observations into actual set of values for the concept.
Concept a have a type in: boolean, numeric, multiple\_selection, single\_selection, and string.
If the concept type is one of the selection types, you must create a set of valid values and coherent among themselves for the concept. You can fill missing possible values from observations.
If the concept type is boolean, numeric or string, you must simply return concept and type. Be careful, you must not consider numeric classification systems like pain, orientation, etc. These are single selection type.
You must return a JSON object with concept (required), type (required) and values (optional).

CONCEPT

\%CONCEPT\%

OBSERVATIONS

\%OBSERVATIONS\%

OUTPUT
\end{promptbox}
\vspace{-0.1in}
\begin{promptbox}[Step 3 - Concept List Expansion]
\footnotesize
Here's a list of concepts in clinical documents. 

\% CONCEPTS \%

Are there missing concepts?
\end{promptbox}
\vspace{-0.1in}
\begin{promptbox}[Step 4 - Concept Combinations]
\scriptsize
You are top clinical semantic expert in medical and clinical knowledge, very precise and very thorough in its work.
You will receive a set of possible clinical concepts in a JSON array.
Your job is to create a JSON object as output containing a rationale and a concept\_list.
The rationale is a description of a realistic clinical case about a patient.
The concept\_list is a JSON array which represents a realistic patient case of a combination of concepts which would be contained a note coming from a nurse's complete speech transcription.
You must provide from a few up to several concepts with "concept" and "value"  as keys, which are concept along their actual observed value respectively.
You must be creative, making unique combinations while maintaining a strong rigor by focusing solely on giving set of concepts.

EXAMPLE OF OUTPUT

\{

    "rationale": "1) This clinical case brings together a realistic and medically coherent set of findings that reflect a plausible presentation of nephrolithiasis with associated complications. The inclusion of right flank pain rated 7 out of 10, described as stabbing, aligns well with typical symptoms of renal colic, particularly from a 2 mm urinary stone, ...",

    "concept\_list": [

        \{"concept": "Pain severity", "value": "7 out of 10"\},
        
        \{"concept": "Pain description", "value": "stabbing pain"\},
        
        \{"concept": "Urinary stone", "value": "True"\}

        ...

    ]

\}

SET OF CONCEPTS

\%CONCEPT\_SET\%

OUTPUT
\end{promptbox}

\begin{promptbox}[Step 5 - Nurse Dictation Generation]
\small
You are a professional medical writer which is an expert in clinical documents and transcripts with deep understanding of nurse speaking patterns and mindset.
I will give you five examples of nurse speech transcripts as examples of style to consider in your creation.
I will also provide you a list of concepts that are present in the speech transcript.
You must output in plain text a highly realistic nurse speech transcript which is a dictation about a patient state.
You must comply with the concept list and include all of the concepts in natural speech as mentions by the nurse.
You must from time to time introduce realistic speech patterns like hesitations, speech phrasing, live revisions of words, etc.
To keep the realism, you must avoid long, unnatural explanations if not needed since the speech transcripts are notes, but you can introduce natural and necessary ones.

EXAMPLE NURSE SPEECH TRANSCRIPT

\% TRANSCRIPT \%

CONCEPT LIST IN SPEECH

\% CONCEPT LIST \%

OUTPUT
\end{promptbox}

\subsection{SYNUR LLM-as-a-Judge Validation}
\label{sec:synur_llmasajudge}

\begin{promptbox}[SYNUR LLM-as-a-Judge Prompt]
\small
You are an expert at medical EHR Flowsheet data analysis who answers in strict, parsable JSON.

You will be given a transcript that a nurse could have created.

Your task is to give a critical assessment of the transcript as a nested JSON object.

For each criteria as a key, the value is a JSON object containing a "rationale" along a "score" on a scale of 1 to 4.

The criteria are: quality and realism.

You will be provided with instructions and examples to complete the task.

Instructions:

Quality

1: The transcript is very low quality and is incoherent. The intended clinical content cannnot be inferred.

2: The transcript is of medium quality and is somewhat coherent. The intended clinical content can be partially inferred.

3: The transcript is high quality and is generally coherent. The intended clinical content can be mostly inferred.

4: The transcript is of very high quality, high clarity and is fully coherent. The intended clinical content can be inferred.

Realism

1: The transcript is not realistic, presenting a situation that is not feasible in a real-world context.

2: The transcript is somewhat realistic, but some aspects may be overly simplified or neglect real-world constraints.

3: The transcript is realistic, but one or two minor aspects may not fully account for practical nuances.

4: The transcript is entirely realistic and likely to be created by a nurse. The style mirrors typical nursing flowsheet documentation.

The following Example Transcripts are real transcript that were provided by nurses.
Example Transcripts:

\%EXAMPLE\_TRANSCRIPTS\%

REMINDER:

The JSON answer should ALWAYS have the following format.

OUTPUT FORMAT:

\{

  "quality": \{"rationale": "string", "score": "integer"\},

  "realism": \{"rationale": "string", "score": "integer"\}

\}

This is really important for Healthcare worker, please do your best to accomplish this task!

Transcript:

\%TRANSCRIPT\%

Now it's time to complete the task.

Remember valid scores are: 1, 2, 3 and 4.

OUTPUT:
\end{promptbox}

\subsection{Aggregation Method of Evaluation Metrics used for SIMORD}
\label{sec:metric}
We evaluate the \textbf{Description}, \textbf{Reason}, \textbf{Type}, and \textbf{Provenance} metrics at the order level. To do this, we first align reference orders with predicted orders at the transcript level maximizing the unigram overlap. Any reference or predicted order that does not find a match is assigned a generic \textit{missing order}. Finally, we combine all matched reference–prediction pairs into a single list and compute precision and recall, treating the \textit{missing orders} as errors. We formalize this computation in the equations below.

Let $O_{\mathrm{ref}}$ and $O_{\mathrm{pred}}$ be the sets of reference and predicted orders.
Let $M \subseteq O_{\mathrm{ref}} \times O_{\mathrm{pred}}$ be the set of matched order pairs with $|M| = TP_o$.
Let $U_p$ be unmatched predicted orders ($FP_o = |U_p|$) and $U_r$ unmatched reference orders ($FN_o = |U_r|$).

Let's first consider the \textbf{Description} and \textbf{Reason} based on ROUGE-1 \cite{lin-2004-rouge}, but the same derivation is also valid for both \textbf{Type} and \textbf{Provenance} metrics. For each matched pair $m \in M$, define the unigram precision and recall on the corresponding sentence pair:

\begin{equation}
\small P_m=\frac{TP_m}{TP_m+FP_m},\ R_m=\frac{TP_m}{TP_m+FN_m}
\end{equation}

and the F1 score is defined as

\begin{equation}
    \small F1_m=\frac{2P_mR_m}{P_m+R_m}.
\end{equation}

Define the aggregated precision and recall as

\begin{equation}
\small
    P_o = \frac{\sum_{m\in M} P_m}{TP_o+FP_o} = \frac{\sum_{m\in M} P_m}{|M|} \cdot \frac{TP_o}{TP_o+FP_o}
\end{equation}

\begin{equation}
\small
    R_o = \frac{\sum_{m\in M} R_m}{TP_o+FN_o} = \frac{\sum_{m\in M} R_m}{|M|} \cdot \frac{TP_o}{TP_o+FN_o}
\end{equation}

considering $ |M|=TP_o $. The second formula highlights that our metrics are also the product of two terms: average of sentence-level precision (or recall) of matched orders, and the match precision (or recall) computed from \textit{missing order}. It follows that

\begin{equation}
    F1_o = \frac{2P_oR_o}{P_o+R_o}
\end{equation}

Since $ 0 \le P_m \le 1 $, we conclude that
$ \sum_{m\in M} P_m \le |M| $, and similarly $ \sum_{m\in M} R_m \le |M| $. Therefore,

\begin{equation}
\small
    P_o \le \frac{TP_o}{TP_o+FP_o},\ R_o \le \frac{TP_o}{TP_o+FN_o}
\end{equation}

which demonstrates that our match scores are the upper bounds of our aggregated metrics. The bound for $F1_o$ follows by the monotonicity of the harmonic mean.

\subsection{Icons in Figures}
\label{sec:icon}
Icons included in Figures \ref{fig:diagram}, \ref{fig:nurse_gen} and \ref{fig:nurse_pipe} are from \textit{IconFinder} (\url{https://www.iconfinder.com/}) without any modification. Here are the creators' attributions:

\begin{itemize}
    \item \textit{Nurse} icon: \textit{BZZRICON}
    \item \textit{Doctor} icon: \textit{iconify}
    \item \textit{Patient} icon: \textit{KonKapp}
    \item \textit{Hospital bed} icon: \textit{Fauzicon}
    \item \textit{LLM} icon: \textit{Eucalyp Studio}
\end{itemize}

\end{document}